\documentclass[10pt,twocolumn,letterpaper]{article}

\usepackage{iccv}
\usepackage{times}
\usepackage{epsfig}
\usepackage{graphicx}
\usepackage{amsmath}
\usepackage{amssymb}
\usepackage{booktabs}
\usepackage{tabulary}
\usepackage{multirow}
\usepackage{overpic}
\usepackage[dvipsnames]{xcolor}
\usepackage[caption=false]{subfig}
\usepackage{xfrac}
\usepackage[british,english,american]{babel}
\usepackage{soul}
\usepackage{xspace}\xspace
\usepackage{adjustbox}
\usepackage{nicefrac}

\definecolor{citecolor}{RGB}{34,139,34}
\usepackage[pagebackref=true,breaklinks=true,letterpaper=true,colorlinks,citecolor=citecolor,bookmarks=false]{hyperref}

\newcommand{\bd}[1]{\textbf{#1}}

\newcommand{\app}{\raise.17ex\hbox{$\scriptstyle\sim$}}
\newcommand{\eq}{{\mkern 0mu=\mkern 0mu}}

\newcolumntype{x}[1]{>{\centering\arraybackslash}p{#1pt}}

\newlength\savewidth\newcommand\shline{\noalign{\global\savewidth\arrayrulewidth
  \global\arrayrulewidth 1pt}\hline\noalign{\global\arrayrulewidth\savewidth}}
\newcommand{\tablestyle}[2]{\setlength{\tabcolsep}{#1}\renewcommand{\arraystretch}{#2}\centering\footnotesize}

\makeatletter\renewcommand\paragraph{\@startsection{paragraph}{4}{\z@}
  {.5em \@plus1ex \@minus.2ex}{-.5em}{\normalfont\normalsize\bfseries}}\makeatother

\def\x{\times}

\newcommand{\m}{$\times$}

\newcommand{\gen}{g}
\newcommand{\paramset}{\Theta}
\newcommand{\aparam}{\theta}
\newcommand{\netset}{\mathcal{N}}
\newcommand{\anet}{n}
\newcommand{\seed}{s}

\newcommand{\ER}{Erd\H{o}s-R\'enyi\xspace}
\newcommand{\BA}{Barab\'asi-Albert\xspace}
\newcommand{\WS}{Watts-Strogatz\xspace}

\def\iccvPaperID{****}\iccvfinalcopy
\ificcvfinal\fi

\begin{document}

\title{\vspace{-.5em} Exploring Randomly Wired Neural Networks for Image Recognition \vspace{-.5em}}

\author{
 Saining Xie \quad Alexander Kirillov \quad Ross Girshick \quad Kaiming He \vspace{3mm}\\
 Facebook AI Research (FAIR) \vspace{-2mm}
}
\maketitle


\begin{abstract}
\vspace{-.5em}
Neural networks for image recognition have evolved through extensive manual design from simple chain-like models to structures with multiple wiring paths. The success of ResNets~\cite{He2016} and DenseNets~\cite{Huang2017} is due in large part to their innovative wiring plans. Now, neural architecture search (NAS) studies are exploring the joint optimization of wiring and operation types, however, the space of possible wirings is constrained and still driven by manual design despite being searched.
In this paper, we explore a more diverse set of connectivity patterns through the lens of \emph{randomly wired neural networks}. To do this, we first define the concept of a \emph{stochastic network generator} that encapsulates the entire network generation process. Encapsulation provides a unified view of NAS and randomly wired networks. Then, we use three classical random graph models to generate randomly wired graphs for networks. The results are surprising: several variants of these random generators yield network instances that have competitive accuracy on the ImageNet benchmark. These results suggest that new efforts focusing on designing better network generators may lead to new breakthroughs by exploring less constrained search spaces with more room for novel design.
\end{abstract}\vspace{-2em}

\section{Introduction}

What we call deep learning today descends from the \emph{connectionist} approach to cognitive science~\cite{Rumelhart1986a,Fodor1988}---a paradigm reflecting the hypothesis that \emph{how computational networks are wired} is crucial for building intelligent machines. Echoing this perspective, recent advances in computer vision have been driven by moving from models with chain-like wiring \cite{Krizhevsky2012,Zeiler2014,Sermanet2014,Simonyan2015} to more elaborate connectivity patterns, \eg, ResNet \cite{He2016} and DenseNet \cite{Huang2017}, that are effective in large part because of how they are wired.

Advancing this trend, neural architecture search (NAS) \cite{Zoph2017,Zoph2018} has emerged as a promising direction for jointly searching wiring patterns and which operations to perform. NAS methods focus on \emph{search} \cite{Zoph2017,Zoph2018,Pham2018,Liu2018a,Luo2018,Liu2019} while implicitly relying on an important---yet largely overlooked---component that we call a \emph{network generator} (defined in \S\ref{sec:generator}). The NAS network generator defines a family of possible wiring patterns from which networks are sampled subject to a learnable probability distribution. However, like the wiring patterns in ResNet and DenseNet, the NAS network generator is \emph{hand designed} and the space of allowed wiring patterns is constrained in a small subset of all possible graphs.
Given this perspective, we ask: \emph{What happens if we loosen this constraint and design novel network generators?}

\begin{figure}[t]\centering
\vspace{-1em}
\includegraphics[width=1.0\linewidth]{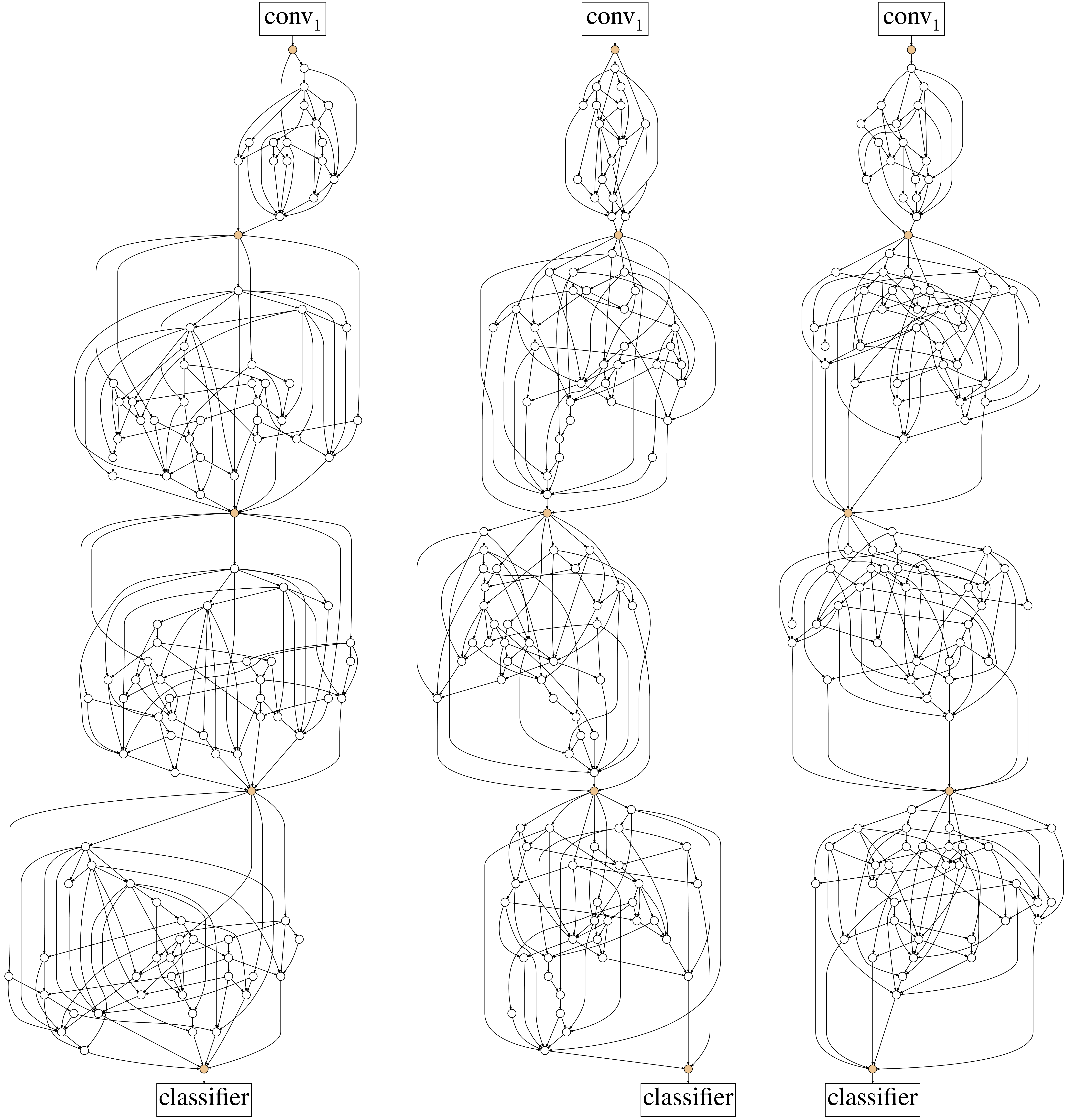}
	\caption{\textbf{Randomly wired neural networks} generated by the classical \WS (WS) \cite{Watts1998} model: these three instances of random networks achieve (left-to-right) 79.1\%, 79.1\%, 79.0\% classification accuracy on ImageNet under a similar computational budget to ResNet-50, which has 77.1\% accuracy.
\label{fig:teaser}}
\vspace{-1em}
\end{figure}

We explore this question through the lens of \emph{randomly wired neural networks} that are sampled from stochastic network generators, in which a human-designed \emph{random process} defines generation.  
To reduce bias from us---the authors of this paper---on the generators, we use \emph{three classical families of random graph models in graph theory} \cite{West1996}: the \ER (ER) \cite{Erdos1960}, \BA (BA) \cite{Albert2002}, and \WS (WS) \cite{Watts1998} models. To define complete networks, we convert a random graph into a directed acyclic graph (DAG) and apply a simple mapping from nodes to their functional roles (\eg, to the same type of convolution).

The results are surprising: several variants of these random generators yield networks with competitive accuracy on ImageNet \cite{Russakovsky2015}. The best generators, which use the WS model, produce multiple networks that outperform or are comparable to their fully manually designed counterparts and the networks found by various neural architecture search methods. We also observe that the variance of accuracy is low for different random networks produced by the same generator, yet there can be clear accuracy gaps between different generators. 
These observations suggest that the network generator \emph{design} is important.

We note that these randomly wired networks are \emph{not} ``prior free'' even though they are random. Many strong priors are in fact implicitly designed into the generator, including the choice of a particular rule and distribution to control the probability of wiring or not wiring certain nodes together.
Each random graph model \cite{Erdos1960,Watts1998,Albert2002} has certain probabilistic behaviors such that sampled graphs likely exhibit certain properties (\eg, WS is highly clustered \cite{Watts1998}).
Ultimately, the generator design determines a probabilistic distribution over networks, and as a result these networks tend to have certain properties. The generator design underlies the prior and thus should not be overlooked.

Our work explores a direction orthogonal to concurrent work on \emph{random search} for NAS \cite{Li2019,Sciuto2019}. These studies show that random search is competitive in ``the NAS search space" \cite{Zoph2017,Zoph2018}, \ie, the ``NAS network generator" in our perspective. Their results can be understood as showing that the prior induced by the NAS generator design tends to produce good models, similar to our observations.
In contrast to \cite{Li2019,Sciuto2019}, our work goes beyond the design of established NAS generators and explores different random generator designs.

Finally, our work suggests a new transition from designing an individual network to \emph{designing a network generator} may be possible, analogous to how our community have transitioned from designing features to designing a network that learns features. Rather than focusing primarily on search with a fixed generator, we suggest designing new network generators that produce new families of models for searching. The importance of the \emph{designed} network generator (in NAS and elsewhere) also implies that machine learning has not been automated (\cf ``AutoML''~\cite{AutoML})---the underlying human design and prior shift from network engineering to network generator engineering.  

\section{Related Work}

\paragraph{Network wiring.} Early recurrent and convolutional neural networks (RNNs and CNNs) \cite{Rumelhart1986,LeCun1989} use chain-like wiring patterns. LSTMs \cite{Hochreiter1997} use more sophisticated wiring to create a gating mechanism. Inception CNNs \cite{Szegedy2015,Szegedy2016a,Szegedy2016} concatenate multiple, irregular branching pathways, while ResNets \cite{He2016} use $x+\mathcal{F}(x)$ as a regular wiring template; DenseNets \cite{Huang2017} use concatenation instead: $[x, \mathcal{F}(x)]$. The LSTM, Inception, ResNet, and DenseNet wiring patterns are effective in general, beyond any individual instantiation.

\paragraph{Neural architecture search (NAS).} Zoph and Le \cite{Zoph2017} define a NAS search space and investigate reinforcement learning (RL) as an optimization algorithm. Recent research on NAS mainly focuses on optimization methods, including RL \cite{Zoph2017,Zoph2018}, progressive \cite{Liu2018a}, gradient-based \cite{Luo2018,Liu2019}, weight-sharing \cite{Pham2018}, evolutionary \cite{Real2018}, and random search \cite{Li2019,Sciuto2019} methods.
The search space in these NAS works, determined by the network generator implicit in \cite{Zoph2017}, is largely unchanged in these works. While this is reasonable for comparing optimization methods, it inherently limits the set of feasible solutions.

\paragraph{Randomly wired machines.} Pioneers of artificial intelligence were originally interested in randomly wired hardware and their implementation in computer programs (\ie, artificial neural networks). In 1940s, Turing \cite{Turing1948} suggested a concept of \emph{unorganized machines}, which is a form of the earliest randomly connected neural networks. One of the first neural network learning machines, designed by Minsky \cite{Minsky1954} in 1950s and implemented using vacuum tubes, was randomly wired. In late 1950s the ``Mark I Perceptron'' visual recognition machine built by Rosenblatt \cite{Rosenblatt1958} used an array of randomly connected photocells.

\paragraph{Relation to neuroscience.} Turing \cite{Turing1948} analogized the unorganized machines to an infant human's cortex.
Rosenblatt \cite{Rosenblatt1958} pointed out that ``\emph{the physical connections of the nervous system ... are not identical from one organism to another}", and ``\emph{at birth, the construction of the most important networks is largely random}." Studies \cite{Watts1998,Varshney2011} have observed that the neural network of a nematode (a worm) with about 300 neurons is a graph with \emph{small-world} properties~\cite{Kochen1989}. Random graph modeling has been used as a tool to study the neural networks of human brains \cite{Bassett2006, Bullmore2009, Bassett2017}.

\paragraph{Random graphs in graph theory.} Random graphs are widely studied in graph theory \cite{West1996}. Random graphs exhibit different probabilistic behaviors depending on the random process defined by the model (\eg, \cite{Erdos1960,Albert2002,Watts1998}). The definition of the random graph model determines the prior knowledge encoded in the resulting graphs (\eg, small-world~\cite{Kochen1989}) and may connect them to naturally occurring phenomena. As a result, random graph models are an effective tool for modeling and analyzing real-world graphs, \eg, social networks, world wide web, citation networks.

\section{Methodology}
\label{sec:method}

We now introduce the concept of a network generator, which is the foundation of randomly wired neural networks.

\subsection{Network Generators}
\label{sec:generator}

We define a \emph{network generator} as a mapping $\gen$ from a parameter space $\paramset$ to a space of neural network architectures $\netset$, $\gen{:}~\paramset \mapsto \netset$. For a given $\aparam \in \paramset$, $\gen(\aparam)$ returns a neural network instance $\anet \in \netset$. The set $\netset$ is typically a family of related networks, for example, VGG nets~\cite{Simonyan2015}, ResNets~\cite{He2016}, or DenseNets~\cite{Huang2017}.

The generator $\gen$ determines, among other concerns, how the computational graph is wired. For example, in ResNets a generator produces a stack of blocks that compute $x+\mathcal{F}(x)$.
The parameters $\aparam$ specify the instantiated network and may contain diverse information.
For example, in a ResNet generator, $\aparam$ can specify the number of stages, number of residual blocks for each stage, depth/width/filter sizes, activation types, \etc.

Intuitively, one may think of $\gen$ as a function in a programming language, \eg Python, that takes a list of arguments (corresponding to $\aparam$), and returns a network architecture.
The network representation $\anet$ returned by the generator is \emph{symbolic}, meaning that it specifies the type of operations that are performed and the flow of data; it does \emph{not} include values of network weights,\footnote{We use \emph{parameters} to refer to network generator arguments and \emph{weights} to refer to the learnable weights and biases of a generated network.} which are learned from data after a network is generated.

\paragraph{Stochastic network generators.} The above network generator $\gen(\theta)$ performs a \emph{deterministic} mapping: given the same $\aparam$, it always returns the same network architecture $\anet$. We can extend $\gen$ to accept an additional argument $\seed$ that is the \emph{seed} of a pseudo-random number generator that is used internally by $g$. Given this seed, one can construct a (pseudo) random family of networks by calling $\gen(\aparam, \seed)$ multiple times, keeping $\aparam$ fixed but changing the value of $\seed = 1, 2, 3, \ldots$. For a fixed value of $\aparam$, a uniform probability distribution over all possible seed values induces a (likely non-uniform) probability distribution over $\netset$. We call generators of the form $\gen(\aparam, \seed)$ \emph{stochastic network generators}.

Before we discuss our method, we provide additional background by reinterpreting the work on NAS~\cite{Zoph2017,Zoph2018} in the context of stochastic network generators.

\paragraph{NAS from the network generator perspective.} The NAS methods of~\cite{Zoph2017,Zoph2018} utilize an LSTM ``controller'' in the process of generating network architectures. But the LSTM is only part of the complete NAS network generator, which is in fact a stochastic network generator, as illustrated next.

The weight matrices of the LSTM are the parameters $\aparam$ of the generator. The output of each LSTM time-step is a probability distribution conditioned on $\aparam$. Given this distribution and the seed $s$, each step samples a construction action (\eg, insert an operator, connect two nodes). The parameters $\aparam$ of the LSTM controller, due to its probabilistic behavior, are optimized (searched for) by RL in \cite{Zoph2017,Zoph2018}.

However, the LSTM is not the only component in the NAS network generator $\gen(\theta, \seed)$. \emph{There are also hand-designed rules defined to map the sampled actions to a computational DAG, and these rules are also part of $\gen$}.
Using the node/edge terminology in graph theory, for a NAS network in \cite{Zoph2018}, if we map a combination operation (\eg, sum) to a node and a unary transformation (\eg, conv) to an edge (see the supplement), the rules of the NAS generator include, but are not limit to:
\begin{itemize}\setlength\itemsep{0.0em}
\item A subgraph to be searched, called a cell \cite{Zoph2018}, always accepts the activations of the output nodes from the 2 immediately preceding cells;  
\item Each cell contains 5 nodes that are wired to 2 and only 2 existing nodes, chosen by sampling from the probability distribution output by the LSTM;
\item All nodes that have no output in a cell are concatenated by an extra node to form a valid DAG for the cell.
\end{itemize}
All of the generation rules, together with the choice of using an LSTM, and other hyper-parameters of the system (\eg, the number of nodes, say, 5), comprise the NAS network generator that produces a full DAG. It is also worth noticing that the view of ``node as combination and edge as transformation" is not the only way to interpret a neural network as a graph, and so it is not the only way to turn a general graph into a neural network (we use a different mapping in \S\ref{sec:randomly_wired_net}).

Encapsulating the \emph{complete} generation process, as we have illustrated, reveals which components are optimized and which are hard-coded. It now becomes explicit that the network space $\netset$ has been carefully restricted by hand-designed rules. For example, the rules listed above suggest that each of the 5 nodes in a cell always \emph{has precisely input degree}\footnote{In graph theory, ``degree" is the number of edges connected to a node. We refer to ``input/output degree" as that of input/output edges to a node.}\emph{ 2 and output degree 1} (see the supplement).
This does not cover all possible 5-(internal-)node graphs. It is in a highly restricted network space.
Viewing NAS from the perspective of a network generator helps explain the recently demonstrated ineffectiveness of sophisticated optimization \vs random search \cite{Li2019,Sciuto2019}: the manual design in the NAS network generator is a \emph{strong} prior, which represents a meta-optimization beyond the search over $\aparam$ (by RL, \eg) and $\seed$ (by random search).

\subsection{Randomly Wired Neural Networks}
\label{sec:randomly_wired_net}

Our analysis of NAS reveals that the network generator is hand-designed and encodes a prior from human knowledge.
It is likely that the design of the network generator plays a considerable role---if so, current methods are short of achieving ``AutoML"~\cite{AutoML} and still involve significant human effort (\cf ``Our experiments show that Neural Architecture Search can design good models from \emph{scratch}.'' \cite{Zoph2017}, emphasis added).
To investigate how important the generator design is, it is not sufficient to compare different optimizers (sophisticated or random) for the same NAS generator; it is necessary to study \emph{new} network generators that are substantially different from the NAS generator.

This leads to our exploration of \emph{randomly wired neural networks}. That is, we will define network generators that yield networks with random graphs, subject to 
different human-specific priors. To minimize the human bias from us---the authors of this paper---on the prior, we will use three \emph{classical random graph models} in our study  (\cite{Erdos1960,Albert2002,Watts1998}; \S\ref{sec:random_graph_models}).
Our methodology for generating randomly wired networks involves the following concepts:

\paragraph{Generating general graphs.} Our network generator starts by generating a general graph (in the sense of graph theory).
It generates a set of nodes and edges that connect nodes, without restricting how the graphs correspond to neural networks. This allows us to freely use any general graph generator from graph theory (ER/BA/WS). 
Once a graph is obtained, it is mapped to a computable neural network.

The mapping from a general graph to neural network operations is in itself arbitrary, and thus also human-designed. We intentionally use a simple mapping, discussed next, so that we can focus on graph wiring patterns.

\paragraph{Edge operations.}
Assuming by construction that the graph is directed, we define that edges are data flow, \ie, a directed edge sends data (a tensor) from one node to another node. 

\paragraph{Node operations.} A node in a directed graph may have some input edges and some output edges. We define the operations represented by one node (Figure~\ref{fig:node_op}) as:

\begin{figure}[t]
\adjustbox{valign=t}{%
\begin{minipage}[c]{0.35\linewidth}
\centering
\includegraphics[width=1.\linewidth]{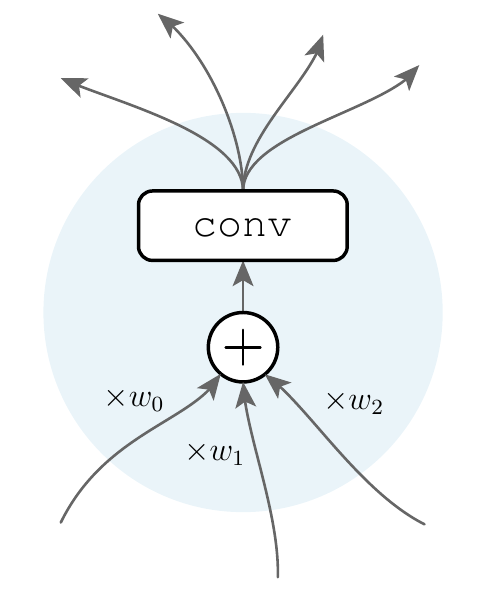}
\end{minipage}\hfill
}
\adjustbox{valign=t}{%
\begin{minipage}[c]{0.61\linewidth}
\caption{\textbf{Node operations} designed for our random graphs. Here we illustrate a node (blue circle) with 3 input edges and 4 output edges. The aggregation is done by weighted sum with learnable positive weights $w_0$, $w_1$, $w_2$. The transformation is a ReLU-convolution-BN triplet, simply denoted as \texttt{conv}. The transformed data are sent out as 4 copies.
}
\label{fig:node_op}
\end{minipage}
}
\vspace{-1.5em}
\end{figure}

\vspace{.3em}
\noindent ~-~\emph{Aggregation}: The input data (from one or more edges) to a node are combined via a weighted sum; the weights are learnable and positive.\footnote{Applying sigmoid on unrestricted weights ensures they are positive.}

\vspace{.5em}
\noindent ~-~\emph{Transformation}: The aggregated data is processed by a transformation defined as a ReLU-convolution-BN triplet\footnote{Instead of a triplet with a convolution followed by BN \cite{Ioffe2015} then ReLU \cite{Nair2010}, we use the ReLU-convolution-BN triplet, as it means the aggregation (at the next nodes) can receive positive and negative activation, preventing the aggregated activation from being inflated in case of a large input degree.} \cite{He2016a}. The same type of convolution is used for all nodes, \eg, a 3$\times$3 separable convolution\footnote{Various implementations of separable convolutions exist. We use the form of \cite{Chollet2017}: a 3$\times$3 separable convolution is a 3$\times$3 \emph{depth-wise} convolution 
followed by a 1$\x$1 convolution, with no non-linearity in between.
} by default.

\vspace{.5em}
\noindent ~-~\emph{Distribution}: The same copy of the transformed data is sent out by the output edges of the node.

\vspace{.5em}

These operations have some nice properties:

\vspace{.3em}
(i)~Additive aggregation (unlike concatenation) maintains the same number of output channels as input channels, and this prevents the convolution that follows from growing large in computation, which may increase the importance of nodes with large input degree simply because they increase computation, not because of how they are wired.

\vspace{.3em}
(ii)~The transformation should have the same number of output and input channels (unless switching stages; discussed later), to make sure the transformed data can be combined with the data from any other nodes. Fixing the channel count then keeps the FLOPs (floating-point operations) and parameter count unchanged for each node, regardless of its input and output degrees.

\vspace{.3em}
(iii)~Aggregation and distribution are almost parameter-free (except for a negligible number of parameters for weighted summation), regardless of input and output degrees. Also, given that every edge is parameter-free the overall FLOPs and parameter count of a graph are roughly proportional to the number of nodes, and nearly independent of the number of edges.

\vspace{.5em}
These properties nearly decouple FLOPs and parameter count from network wiring, \eg, the deviation of FLOPs is typically $\pm$2\% among our random network instances or different generators.
This enables the comparison of different graphs without inflating/deflating model complexity. Differences in task performance are therefore reflective of the properties of the wiring pattern. 

\paragraph{Input and output nodes.} Thus far, a general graph is not yet a valid neural network even given the edge/node operations, because it may have multiple input nodes (\ie, those without any input edge) and multiple output nodes. It is desirable to have a single input and a single output for typical neural networks, \eg, for image classification. We apply a simple post-processing step.

For a given general graph, we create a single extra node that is connected to all original input nodes. This is the unique input node that sends out the same copy of input data to all original input nodes. Similarly, we create a single extra node that is connected to all original output nodes. This is the unique output node; we have it compute the (unweighted) average from all original output nodes.
These two nodes perform no convolution.
When referring to the node count $N$, we exclude these two nodes.

\paragraph{Stages.} With unique input and output nodes, it is sufficient for a graph to represent a valid neural network. Nevertheless, in image classification in particular, networks that maintain the full input resolution throughout are not desirable. It is common \cite{Krizhevsky2012,Simonyan2015,He2016,Zoph2018} to divide a network into \emph{stages} that progressively down-sample feature maps.

We use a simple strategy: the random graph generated above defines one stage. Analogous to the stages in 
a ResNet, \eg, conv$_{1,2,3,4,5}$ \cite{He2016}, our entire network consists of multiple stages. 
One random graph represents one stage, and it is connected to its preceding/succeeding stage by its unique input/output node. For all nodes that are directly connected to the input node, their transformations are modified to have a stride of 2. The channel count in a random graph is increased by 2$\times$ when going from one stage to the next stage, following \cite{He2016}.

Table~\ref{tab:arch} summarizes the randomly wired neural networks, referred to as \textbf{\emph{RandWire}}, used in our experiments. They come in small and regular complexity regimes (more in \S\ref{sec:exp}). For conv$_1$ and/or conv$_2$ we use a single convolutional layer for simplicity with multiple random graphs following. The network ends with a classifier output (Table~\ref{tab:arch}, last row). Figure~\ref{fig:teaser} shows full computation graphs of three randomly wired network samples.

\newcommand{\graph}{\emph{\textbf{random wiring}}}
\newcommand{\conv}{\texttt{conv}}
\newcolumntype{x}[1]{>\centering p{#1pt}}
\newcommand{\ft}[1]{\fontsize{#1pt}{1em}\selectfont}
\renewcommand\arraystretch{1.1}
\setlength{\tabcolsep}{6pt}
\begin{table}[t]
\begin{center}
\footnotesize
\begin{tabular}{c|c|x{60}|c}
 stage & output & \emph{small regime} & \emph{{regular regime}} \\
\shline
\multirow{1}{*}{$\text{conv}_1$} & \multirow{1}{*}{\ft{7} 112\m112}
&  \multicolumn{2}{c}{\ft{7} 3\m3 \conv, $C/2$} \\
\hline
\multirow{2}{*}{$\text{conv}_2$} & \multirow{2}{*}{\ft{7} 56\m56}
  & \multirow{2}{*}{3\m3 \conv, $C$} & \graph\\
  & & & $N/2$, $C$ \\
\hline
\multirow{2}{*}{$\text{conv}_3$} & \multirow{2}{*}{\ft{7} 28\m28}
& \graph & \graph \\
& & {$N$, $C$} & {$N$, $2C$} \\
\hline
\multirow{2}{*}{$\text{conv}_4$} & \multirow{2}{*}{\ft{7} 14\m14}
& \graph & \graph \\
& & {$N$, $2C$} & {$N$, $4C$} \\
\hline
\multirow{2}{*}{$\text{conv}_5$} & \multirow{2}{*}{\ft{7} 7\m7}
& \graph & \graph \\
& & {$N$, $4C$} & {$N$, $8C$} \\
\hline
\multirow{2}{*}{$\text{classifier}$} & \multirow{2}{*}{\ft{7} 1$\times$1}
& \multicolumn{2}{c}{\ft{7} 1\m1 \conv, 1280-d} \vspace{-.2em} \\
& & \multicolumn{2}{c}{\ft{7} global average pool, 1000-d \emph{fc}, softmax} \\
\end{tabular}
\end{center}
\vspace{-.5em}
\caption{
\textbf{RandWire architectures} for small and regular computation networks.
A random graph is denoted by the node count ($N$) and channel count for each node ($C$).
We use \texttt{conv} to denote a ReLU-Conv-BN triplet (expect conv$_1$ is Conv-BN).
The input size is 224$\times$224 pixels. The change of the output size implies a stride of 2 (omitted in table) in the convolutions that are right after the input of each stage.
}
\label{tab:arch}
\vspace{-.5em}
\end{table}

\subsection{Random Graph Models}
\label{sec:random_graph_models}

We now describe in brief the three classical random graph models used in our study. We emphasize that these random graph models are not proposed by this paper; we describe them for completeness.
The three classical models all generate undirected graphs; we use a simple heuristic to turn them into DAGs (see the supplement).

\paragraph{\ER (ER).} 
In the ER model~\cite{Gilbert1959,Erdos1960}, with $N$ nodes, an edge between two nodes is connected with probability $P$, independent of all other nodes and edges.
This process is iterated for all pairs of nodes. The ER generation model has only a single parameter $P$, and is denoted as ER$(P)$.

Any graph with $N$ nodes has non-zero probability of being generated by the ER model, including graphs that are disconnected.
However, a graph generated by ER$(P)$ has high probability of being a single connected component if $P > \frac{\ln(N)}{N}$~\cite{Erdos1960}. This provides one example of an implicit bias introduced by a generator.

\paragraph{\BA (BA).} 
The BA model~\cite{Albert2002} generates a random graph by sequentially adding new nodes.
The initial state is $M$ nodes without any edges ($1 \leq M < N$). 
The method sequentially adds a new node with $M$ new edges.
For a node to be added, it will be connected to an existing node $v$ with probability proportional to $v$'s degree. The new node repeatedly adds non-duplicate edges in this way until it has $M$ edges. Then this is iterated until the graph has $N$ nodes. The BA generation model has only a single parameter $M$, and is denoted as BA$(M)$.

Any graph generated by BA$(M)$ has exactly $M$$\cdot$$(N$$-$$M)$ edges. 
So the set of all graphs generated by BA$(M)$ is a \emph{subset} of all possible $N$-node graphs---this gives one example on how an underlying prior can be introduced by the graph generator in spite of randomness. 

\paragraph{\WS (WS).}
The WS model~\cite{Watts1998} was defined to generate small-world graphs~\cite{Kochen1989}. Initially, the $N$ nodes are regularly placed in a ring and each node is connected to its $K/2$ neighbors on both sides ($K$ is an even number).
Then, in a clockwise loop, for every node $v$, the edge that connects $v$ to its clockwise $i$-th next node is \emph{rewired} with probability $P$. ``Rewiring" is defined as uniformly choosing a random node that is not $v$ and that is not a duplicate edge. This loop is repeated $K/2$ times for $1$$\leq$$i$$\leq$$K/2$.
$K$ and $P$ are the only two parameters of the WS model, denoted as WS$(K, P)$.

Any graph generated by WS$(K, P)$ has exactly $N$$\cdot$$K$ edges. 
WS$(K, P)$ only covers a small subset of all possible $N$-node graphs too, but this subset is different from the subset covered by BA. This provides an example on how a different underlying prior has been introduced.

\begin{figure*}[t]\centering
\vspace{-1em}
\includegraphics[width=.9\linewidth]{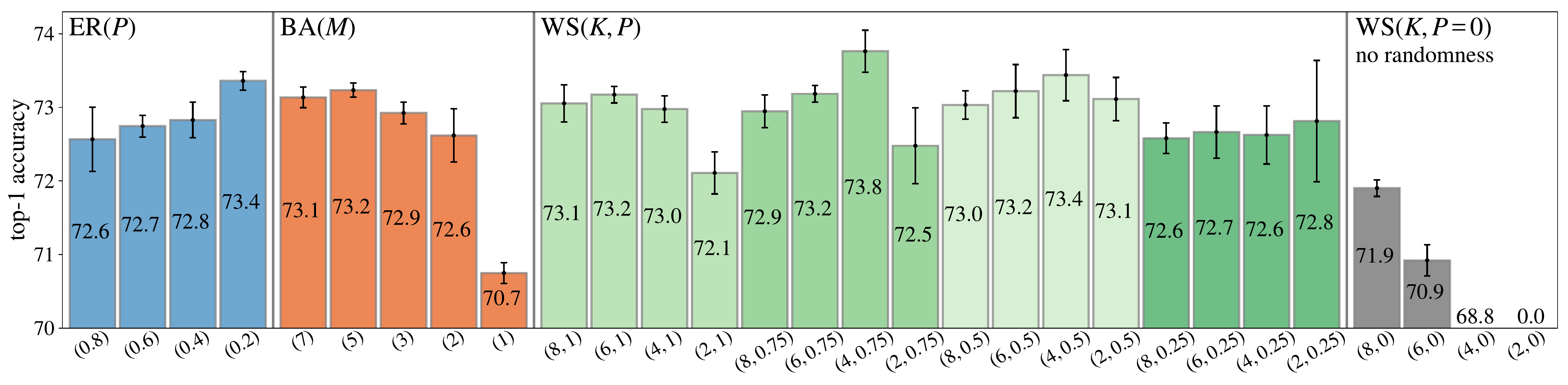}
\vspace{-.3em}
\caption{\bd{Comparison on random graph generators: ER, BA, and WS} in the small computation regime. Each bar represents the results of a generator under a parameter setting for $P$, $M$, or $(K, P)$ (tagged in x-axis).
The results are ImageNet top-1 accuracy, shown as mean and standard deviation (std) over 5 random network instances sampled by a generator.
At the rightmost, WS$(K, P\eq0)$ has no randomness.
\label{fig:sweep-and-graphs}}
\vspace{-1em}
\end{figure*}

\begin{figure*}[t]\centering
\includegraphics[width=.95\linewidth]{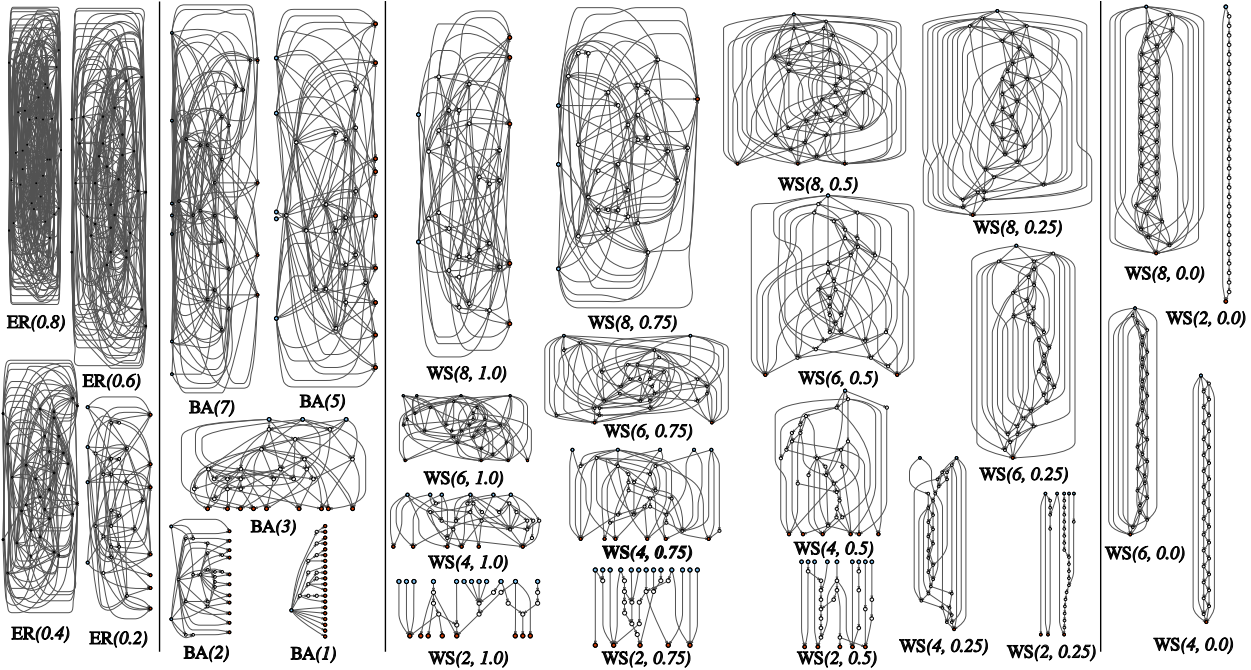}
\caption{\textbf{Visualization of the random graphs generated by ER, BA, and WS}. Each plot represents one random graph instance sampled by the specified generator. The generators are those in Figure~\ref{fig:sweep-and-graphs}.
The node count is $N\eq32$ for each graph. A blue/red node denotes an input/output node, to which an extra unique input/output node (not shown) will be added (see \S\ref{sec:randomly_wired_net}).
\label{fig:graphs}}
\vspace{-1em}
\end{figure*}


\subsection{Design and Optimization}
\label{sec:opt}

Our randomly wired neural networks are generated by a stochastic network generator $\gen(\aparam, \seed)$.
The random graph parameters, namely, $P$, $M$, $(K, P)$ in ER, BA, WS respectively, are part of the parameters $\aparam$. 
The ``optimization" of such a 1- or 2-parameter space is essentially done by \emph{trial-and-error} by human designers, \eg, by line/grid search. Conceptually, such ``optimization" is not distinct from many other \emph{designs} involved in our and other models (including NAS), \eg, the number of nodes, stages, and filters.

Optimization can also be done by scanning the random seed $\seed$, which is an implementation of random search. Random search is possible for any stochastic network generator, including ours and NAS. But as we present by experiment, the accuracy variation of our networks is small for different seeds $\seed$, suggesting that the benefit of random search may be small. So we perform \emph{no random search} and report \emph{mean} accuracy of multiple random network instances.
As such, our network generator has \emph{minimal optimization} (1- or 2-parameter grid search) beyond their hand-coded \emph{design}.

\section{Experiments}
\label{sec:exp}

We conduct experiments on the ImageNet 1000-class classification task \cite{Russakovsky2015}. We train on the training set with $\app$1.28M images and test on the 50K validation images.

\paragraph{Architecture details.} 
Our experiments span a small computation regime (\eg, MobileNet \cite{Howard2017} and ShuffleNet \cite{Zhang2018}) and a regular computation regime (\eg, ResNet-50/101 \cite{He2016}).
RandWire nets in these regimes are in Table~\ref{tab:arch}, where $N$ nodes and $C$ channels determine network complexity.
We set $N\eq32$, and then set $C$ to the {nearest integer} such that target model complexity is met: $C\eq78$ in the small regime, and $C\eq109$ or $154$ in the regular regime.

\paragraph{Random seeds.} For each generator, we randomly sample 5 network instances (5 random seeds), train them from scratch, and evaluate accuracy for each instance. 
To emphasize that we perform \emph{no random search} for each generator, we report the classification accuracy with ``{mean$\pm$std}'' for \emph{all} 5 random seeds (\ie, we do \emph{not} pick the best).
We use the same seeds 1, $\ldots$, 5 for all experiments.

\paragraph{Implementation details.} 
We train our networks for 100 epochs, unless noted. We use a half-period-cosine shaped learning rate decay \cite{Loshchilov2016,Huang2017}. The initial learning rate is 0.1, the weight decay is 5e-5, and the momentum is 0.9. We use label smoothing regularization \cite{Szegedy2016} with a coefficient of 0.1.
Other details of the training procedure are the same as \cite{Goyal2017}.

\subsection{Analysis Experiments}

\paragraph{Random graph generators.}
Figure~\ref{fig:sweep-and-graphs} compares the results of different generators in the small computation regime: each RandWire net has $\app$580M FLOPs. Figure~\ref{fig:graphs} visualizes one example graph for each generator.
The graph generator is specified by the random graph model (ER/BA/WS) and its set of parameters: \eg, ER$(0.2)$.
We observe:

\emph{All random generators provide decent accuracy} over all 5 random network instances; none of them fails to converge. ER, BA, and WS all have certain settings that yield mean accuracy of $>$73\%, within a $<$1\% gap from the best mean accuracy of 73.8\% from WS$(4, 0.75)$.

Moreover, \emph{the variation among the random network instances is low}. Almost all random generators in Figure~\ref{fig:sweep-and-graphs} have an standard deviation (std) of 0.2$\app$0.4\%. As a comparison, training the same instance of a ResNet-50 multiple times has a typical std of 0.1$\app$0.2\%~\cite{Goyal2017}.
The observed low variance of our random generators suggests that even without random search (\ie, picking the best from several random instances), it is likely that the accuracy of a network instance is close to the mean accuracy, subject to some noise.

On the other hand, different random generators may have a gap between their mean accuracies, \eg, BA$(1)$ has 70.7\% accuracy and is $\app$3\% lower than WS$(4, 0.75)$. This suggests that random generator design, including the wiring priors (BA \vs WS) and generation parameters, plays an important role in the accuracy of sampled network instances. 

Figure~\ref{fig:sweep-and-graphs} also includes a set of \emph{non-random} generators: WS$(K, P\eq0)$. ``$P\eq0$" means no random rewiring. Interestingly, the results of WS$(K, P\eq0)$ are all worse than their WS$(K, P$$>$$0)$ counterparts for any fixed $K$ in Figure~\ref{fig:sweep-and-graphs}.

\paragraph{Graph damage.} We explore \emph{graph damage} by randomly removing one node or edge---an ablative setting inspired by \cite{LeCun1990,Veit2016}. Formally, given a network instance \emph{after training}, we remove one node or one edge from the graph and evaluate the validation accuracy \emph{without any further training}. 

\begin{figure}[t]
\vspace{-.5em}
\centering
\includegraphics[width=1.02\linewidth]{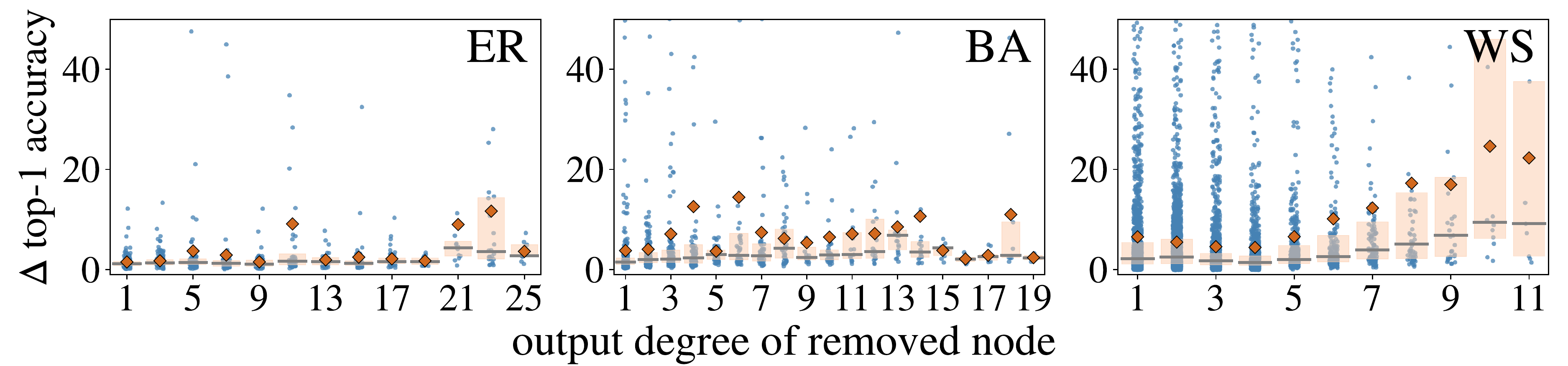}
\includegraphics[width=1.02\linewidth]{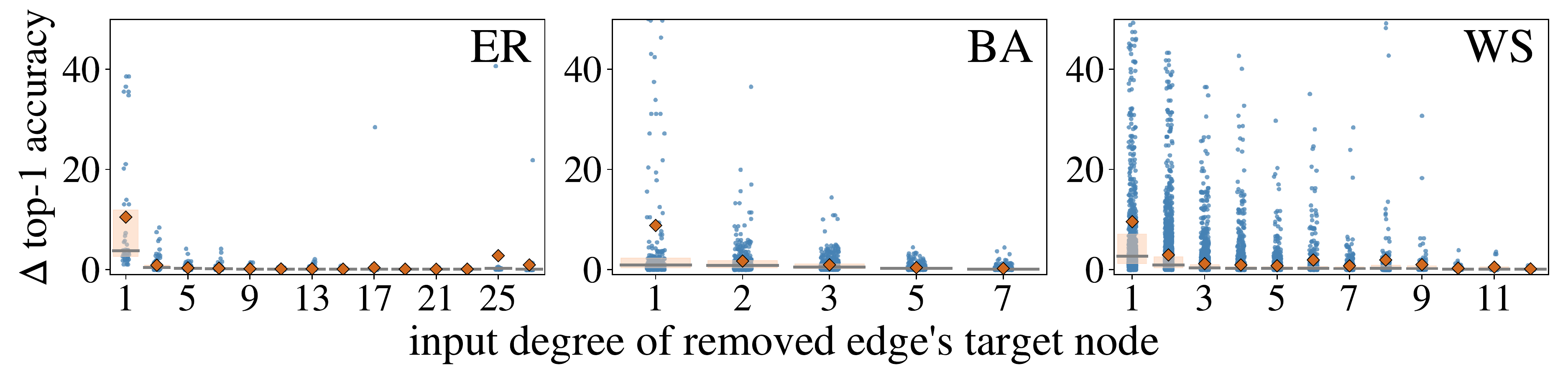}
\caption{
\textbf{Graph damage ablation}. We randomly \textbf{\emph{remove one node}} (top) or \textbf{\emph{remove one edge}} (bottom) from a graph after the network is trained, and evaluate the loss ($\Delta$) in accuracy on ImageNet. From left to right are ER, BA, and WS generators.
Red circle: \emph{mean}; gray bar: \emph{median}; orange box: \emph{interquartile range}; blue dot: \emph{an individual damaged instance}.
\label{fig:element-removal}}
\vspace{-.5em}
\end{figure}

When a node is removed, we evaluate the accuracy loss ($\Delta$) \vs the output degree of that node (Figure~\ref{fig:element-removal}, top). It is clear that ER, BA, and WS behave differently under such damage. For networks generated by WS, the mean degradation of accuracy is larger when the output degree of the removed node is higher. This implies that ``hub" nodes in WS that send information to many nodes are influential.

When an edge is removed, we evaluate the accuracy loss \vs the input degree of this edge's target node (Figure~\ref{fig:element-removal}, bottom). If the input degree of an edge's target node is smaller, removing this edge tends to change a larger portion of the target node's inputs. This trend can be seen by the fact that the accuracy loss is generally decreasing along the x-axis in Figure~\ref{fig:element-removal} (bottom). The ER model is less sensitive to edge removal, possibly because in ER's definition wiring of every edge is independent.

\begin{figure}[t]
\vspace{-.2em}
\centering
\includegraphics[width=.99\linewidth]{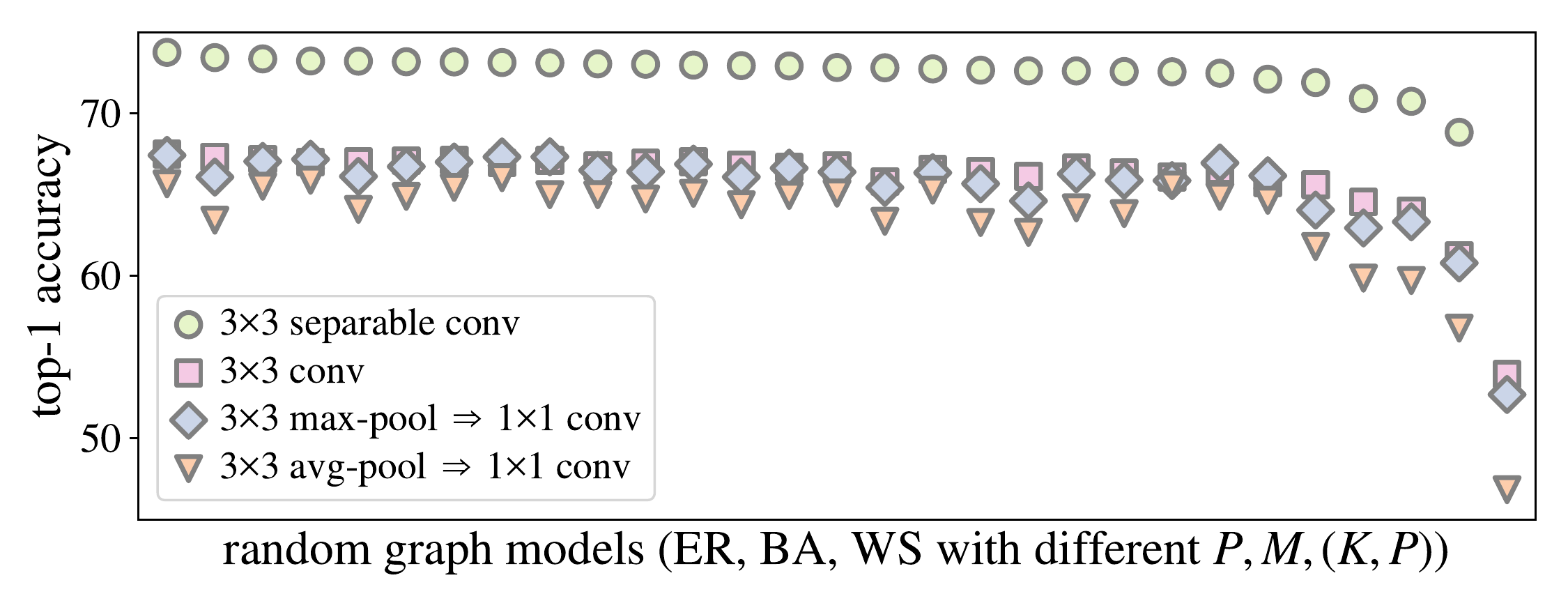}
\vspace{-.5em}
\caption{\textbf{Alternative node operations}.
Each column is the mean accuracy of the same set of 5 random graphs equipped with different node operations, sorted by ``3$\times$3 separable conv" (from Figure~\ref{fig:sweep-and-graphs}). 
The generators roughly maintain their orders of accuracy.
\label{fig:ops-comparison}}
\vspace{-1.5em}
\end{figure}

\paragraph{Node operations.} Thus far, all models in our experiment use a 3$\times$3 separable convolution as the ``\texttt{conv}" in Figure~\ref{fig:node_op}. Next we evaluate alternative choices. We consider: (i) 3$\times$3 (regular) convolution, and (ii) 3$\times$3 max-/average-pooling followed by a 1$\times$1 convolution. We replace the transformation of \emph{all nodes} with the specified alternative.
We adjust the factor $C$ to keep the complexity of all alternative networks.

Figure~\ref{fig:ops-comparison} shows the mean accuracy for each of the generators listed in Figure~\ref{fig:sweep-and-graphs}. Interestingly, almost all networks still converge to non-trivial results. Even ``3$\times$3 pool with 1$\times$1 conv" performs similarly to ``3$\times$3 conv". The network generators roughly maintain their accuracy ranking despite the operation replacement; in fact, the Pearson correlation between any two series in Figure~\ref{fig:element-removal} is $0.91$$\app$$0.98$. This suggests that the network wiring plays a role somewhat orthogonal to the role of the chosen operations.

\subsection{Comparisons}

\newcommand{\pmstd}[1]{$_{\pm\text{#1}}$}
\vspace{-.5em}
\begin{table}[t]\centering
\tablestyle{4.5pt}{1.05}\begin{tabular}{l|ll|ll}
network & top-1 acc. & top-5 acc. & \ft{7}FLOPs (M) & \ft{7}params (M) \\
\shline
MobileNet~\cite{Howard2017} & 70.6 & 89.5 & 569 & 4.2 \\
MobileNet v2~\cite{Sandler2018} & 74.7 & - & 585 & 6.9 \\
ShuffleNet~\cite{Zhang2018} & 73.7 & 91.5 & 524 & 5.4 \\
ShuffleNet v2~\cite{Ma2018} & \textbf{74.9} & 92.2 & 591 & 7.4 \\
\hline
NASNet-A~\cite{Zoph2018} & 74.0 & 91.6 & 564 & 5.3 \\
NASNet-B~\cite{Zoph2018} & 72.8 & 91.3 & 488 & 5.3 \\
NASNet-C~\cite{Zoph2018} & 72.5 & 91.0 & 558 & 4.9 \\
Amoeba-A~\cite{Real2018} & 74.5 & 92.0 & 555 & 5.1 \\
Amoeba-B~\cite{Real2018} & 74.0 & 91.5 & 555 & 5.3 \\
Amoeba-C~\cite{Real2018} & \textbf{75.7} & \textbf{92.4} & 570 & 6.4 \\
PNAS~\cite{Liu2018a} & 74.2 & 91.9 & 588 & 5.1 \\
DARTS~\cite{Liu2019} & 73.1 & 91.0 & 595 & 4.9 \\
\hline
\textbf{RandWire-WS} & \textbf{74.7}\pmstd{0.25} & \textbf{92.2}\pmstd{0.15} & 583\pmstd{6.2} & 5.6\pmstd{0.1}
\end{tabular}
\vspace{.7em}
\caption{\textbf{ImageNet: small computation regime} (\ie, $<$600M FLOPs).
RandWire results are the mean accuracy ($\pm$std) of 5 random network instances, with WS$(4, 0.75)$. Here we train for 250 epochs similar to \cite{Zoph2018,Real2018,Liu2018a,Liu2019}, for fair comparisons.
\label{tab:imagenet-low-compute}
}
\vspace{-.5em}
\end{table}

\paragraph{Small computation regime.} Table~\ref{tab:imagenet-low-compute} compares our results in the \emph{small computation regime}, a common setting studied in existing NAS papers. Instead of training for 100 epochs, here we train for 250 epochs following settings in \cite{Zoph2018,Real2018,Liu2018a,Liu2019} for fair comparisons.

RandWire with WS$(4, 0.75)$ has mean accuracy of 74.7\% (with min 74.4\% and max 75.0\%). This result is better than or comparable to all existing hand-designed wiring (MobileNet/ShuffleNet) and NAS-based results, except for AmoebaNet-C~\cite{Real2018}. The \emph{mean} accuracy achieved by RandWire is a competitive result, especially considering that we perform no random search in our random generators, and that we use a single operation type for all nodes.

\begin{table}[t]\centering
\tablestyle{3pt}{1.05}\begin{tabular}{l|ll|ll}
network & top-1 acc. & top-5 acc. & \ft{7}FLOPs (B) & \ft{7}params (M) \\\shline
ResNet-50~\cite{He2016} & 77.1  & 93.5 & 4.1 & 25.6 \\
ResNeXt-50~\cite{Xie2017} & 78.4  & 94.0 & 4.2 & 25.0 \\
\textbf{RandWire-WS}, $C\eq109$ & \textbf{79.0}\pmstd{0.17}  & \textbf{94.4}\pmstd{0.11} & 4.0\pmstd{0.09} & 31.9\pmstd{0.66} \\
\hline
ResNet-101~\cite{He2016} & 78.8 & 94.4 & 7.8 & 44.6 \\
ResNeXt-101~\cite{Xie2017} & 79.5 & 94.6 & 8.0 & 44.2 \\
\textbf{RandWire-WS}, $C\eq154$ & \textbf{80.1}\pmstd{0.19}  & \textbf{94.8}\pmstd{0.18} & 7.9\pmstd{0.18} & 61.5\pmstd{1.32} \\
\end{tabular}
\vspace{.5em}
\caption{\textbf{ImageNet: regular computation regime} with FLOPs comparable to ResNet-50 (top) and to ResNet-101 (bottom). ResNeXt is the 32$\times$4 version \cite{Xie2017}. RandWire is WS$(4, 0.75)$.
\label{tab:imagenet-regular-compute}
}
\vspace{-0.5em}
\end{table}

\begin{table}[t]\centering
\tablestyle{1.6pt}{1.05}
\begin{tabular}{l|c|c|ll|ll}
\vspace{-.3em}
\multirow{2}{*}{\ft{7}network} & \ft{7}test & \multirow{2}{*}{\ft{7}epochs}  & \multirow{2}{*}{\ft{7}top-1 acc.} & \multirow{2}{*}{\ft{7}top-5 acc.} & \multirow{2}{*}{\ft{7}FLOPs (B)} & \multirow{2}{*}{\ft{7}params (M)} \\
                               & \ft{7}size &                          &  &  &                           &                            \\
\shline
NASNet-A~\cite{Zoph2018} & \ft{7}331$^2$ & \ft{7}$>$250 & 82.7 & 96.2 & 23.8 & 88.9 \\
Amoeba-B~\cite{Real2018} & \ft{7}331$^2$ & \ft{7}$>$250 & 82.3 & 96.1 & 22.3 & 84.0 \\
Amoeba-A~\cite{Real2018} & \ft{7}331$^2$ & \ft{7}$>$250 & 82.8 & 96.1 & 23.1 & 86.7 \\
PNASNet-5~\cite{Liu2018a} & \ft{7}331$^2$ & \ft{7}$>$250 & 82.9 & 96.2 & 25.0 & 86.1 \\
\hline
\textbf{RandWire-WS} & \ft{7}320$^2$ & \ft{7}100 & 81.6\pmstd{0.13} & 95.6\pmstd{0.07} & 16.0\pmstd{0.36} & 61.5\pmstd{1.32} \\
\end{tabular}
\vspace{.5em}
\caption{\textbf{ImageNet: large computation regime}. Our networks are the same as in Table~\ref{tab:imagenet-regular-compute} ($C\eq154$), but we evaluate on 320$\times$320 images instead of 224$\times$224. Ours are only trained for 100 epochs.}
\label{tab:imagenet-compare-nas}
\vspace{-1em}
\end{table}

\paragraph{Regular computation regime.} Next we compare the RandWire networks with ResNet-50/101 \cite{He2016} under similar FLOPs. In this regime, we use a regularization method inspired by our edge removal analysis: for each training mini-batch, we randomly remove one edge whose target node has input degree $>$ 1 with probability of 0.1. This regularization is similar to DropPath adopted in NAS \cite{Zoph2018}. We train with a weight decay of 1e-5 and a DropOut \cite{Hinton2012} rate of 0.2 in the classifier \emph{fc} layer. Other settings are the same as the small computation regime. We train the ResNet/ResNeXt competitors using the recipe of \cite{Goyal2017}, but with the cosine schedule and label smoothing, for fair comparisons.

Table~\ref{tab:imagenet-regular-compute} compares RandWire with ResNet and ResNeXt under similar FLOPs as ResNet-50/101. Our mean accuracies are respectively 1.9\% and 1.3\% higher than ResNet-50 and ResNet-101, and are 0.6\% higher than the ResNeXt counterparts. Both ResNe(X)t and RandWire can be thought of as hand-designed, but ResNe(X)t is based on designed wiring patterns, while RandWire uses a designed stochastic generator. These results illustrate different roles that manual design can play.

\paragraph{Larger computation.} For completeness, we compare with the most accurate NAS-based networks, which use more computation. For simplicity, we use \emph{the same trained networks} as in Table~\ref{tab:imagenet-regular-compute}, but only increase the test image size to 320$\times$320 without retraining. Table~\ref{tab:imagenet-compare-nas} compares the results.

Our networks have mean accuracy 0.7\%$\app$1.3\% lower than the most accurate NAS results, but ours use only $\app$2/3 FLOPs and $\app$3/4 parameters.
Our networks are trained for 100 epochs and not on the target image size, \vs the NAS methods which use $>$250 epochs and train on the target 331$\times$331 size. Our model has no search on operations, unlike NAS. These gaps will be explored in future work.

\begin{table}[t]\centering
\tablestyle{4.5pt}{1.05}\begin{tabular}{l|ccc|ccc}
backbone & AP & AP\textsubscript{50} & AP\textsubscript{75} & AP\textsubscript{S} & AP\textsubscript{M} & AP\textsubscript{L} \\\shline
ResNet-50~\cite{He2016} & 37.1 & 58.8 & 39.7 & 21.9 & 40.8 & 47.6\\
ResNeXt-50~\cite{Xie2017} & 38.2 & 60.5 & 41.3 & 23.0 & 41.5 & 48.8\\
\textbf{RandWire-WS}, $C\eq109$ & \bd{39.9} & \bd{61.9} & \bd{43.3} & \bd{23.6} & \bd{43.5} & \bd{52.7}\\
\hline
ResNet-101~\cite{He2016} & 39.8 & 61.7 & 43.3 & 23.7 & 43.9 & 51.7\\
ResNeXt-101~\cite{Xie2017} & 40.7 & 62.9 & 44.5 & 24.4 & 44.8 & 52.7\\
\textbf{RandWire-WS}, $C\eq154$ & \bd{41.1} & \bd{63.1} & \bd{44.6} & \bd{24.6} & \bd{45.1} & \bd{53.0}\\
\end{tabular}
\vspace{.5em}
\caption{\textbf{COCO object detection} results fine-tuned from the networks in Table~\ref{tab:imagenet-regular-compute}, reported on the \texttt{val2017} set.
The backbone networks have comparable FLOPs to ResNet-50 or ResNet-101.
\label{tab:detection}
}
\vspace{-1em}
\end{table}

\paragraph{COCO object detection.}
Finally, we report the transferability results by fine-tuning the networks for COCO object detection \cite{Lin2014}. We use Faster R-CNN \cite{Ren2015} with FPN \cite{Lin2017} as the object detector. Our fine-tuning is based on 1$\times$ setting of the publicly available \texttt{Detectron} \cite{Detectron2018}. We simply replace the backbones with those in Table~\ref{tab:imagenet-regular-compute} (regular regime).

Table~\ref{tab:detection} compares the object detection results. A trend is observed similar to that in the ImageNet experiments in Table~\ref{tab:imagenet-regular-compute}. These results indicate that the features learned by our randomly wired networks can also transfer. 

\section{Conclusion}

We explored randomly wired neural networks driven by three classical random graph models from graph theory.
The results were surprising: the mean accuracy of these models is competitive with hand-designed \emph{and} optimized models from recent work on neural architecture search.
Our exploration was enabled by the novel concept of a \emph{network generator}.
We hope that future work exploring new generator designs may yield new, powerful networks designs.

\appendix
\section{Appendix}

\begin{figure}[h]\centering
\includegraphics[width=.9\linewidth]{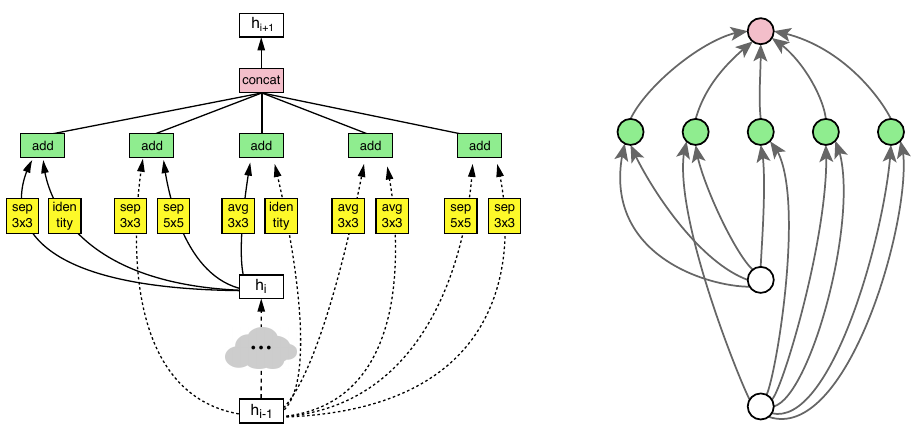}
\caption{Mapping a NAS cell (left, credit: \cite{Zoph2018}) to a graph (right).
}
\label{fig:nas}
\vspace{-.5em}
\end{figure}

\paragraph{Mapping a NAS cell to a graph.}
If one maps a combining op (\eg, addition, concatenation) to a node, and a unary transformation (\eg, 3$\times$3 conv, 5$\times$5 conv, identity) to an edge (Figure~\ref{fig:nas}, right), then all cells in the NAS search space share this property: \emph{internal nodes all have precisely input degree 2 and output degree 1}. This is an implicit prior induced by the design.

The mapping from the NAS cell to a graph is not unique. One may map both combining and unary transformations to nodes, and data flow to edges (Figure~\ref{fig:nas}, left). The above property on the NAS search space can be instead described as: internal \emph{merging} nodes all have precisely input degree 2 and output degree 1.

\paragraph{Converting undirected graphs into DAGs.}
ER, BA, and WS models generate random undirected graphs. We convert them to DAGs using a simple heuristic: we assign indices to all nodes in a graph, and set the direction of every edge as pointing from the smaller-index node to the larger-index one. This heuristic ensures that there is no cycle in the resulted directed graph. The node indexing strategies for the models are ---
ER: indices are assigned in a random order; BA: the initial $M$ nodes are assigned indices $1$ to $M$, and all other nodes are indexed following their order of adding to the graph;  WS: indices are assigned sequentially in the clockwise order.

{\small \bibliographystyle{ieee} \bibliography{randwire}}

\end{document}